\documentclass[11pt]{article}

\usepackage[margin=1in]{geometry}
\usepackage{amsmath,amssymb,amsthm}
\usepackage{graphicx}
\usepackage{pgfplots}
\pgfplotsset{compat=1.18}

\usepackage{booktabs}
\usepackage{bm}
\usepackage{authblk}

\usepackage[numbers]{natbib}
\usepackage{hyperref}
\hypersetup{
    colorlinks=true,
    linkcolor=blue,
    filecolor=magenta,      
    urlcolor=blue,
    citecolor=red,
}

\usepackage[colorinlistoftodos]{todonotes} 
\usepackage{framed}

\title{Lost in the Middle at Birth: An Exact Theory of Transformer Position Bias}
\author{Borun D Chowdhury}
\affil{Meta, London, W1T 1FB, UK}
\affil{\texttt{borundev@gmail.com}}
\date{}    

\begin{document}

\maketitle

\begin{abstract}
The ``Lost in the Middle'' phenomenon---a U-shaped performance curve
where LLMs retrieve well from the beginning and end of a context but
fail in the middle---is widely attributed to learned Softmax artifacts
or the distance-decay of positional encodings like RoPE. This paper
makes a single, precise claim: \emph{the U-shape is already present at
initialization, before any training or positional encoding takes
effect.} It is an inherent geometric property of the causal decoder
with residual connections.

We model multi-layer causal attention as iterated powers of the
Ces\`{a}ro matrix and derive the exact closed-form influence density
in the continuous limit. Causal masking forces a logarithmic divergence
of gradient influence at the start of the prompt (the Primacy Tail),
while residual connections create an isolated $\mathcal{O}(1)$ anchor
at the final token (the Recency Delta). Between these extremes lies a
factorial dead zone of order $\mathcal{O}(1/(H{-}1)!)$, where $H$ is
the network depth, making middle-context retrieval and training structurally
hostile. We validate empirically that untrained Qwen2 and GPT-2
architectures exhibit this U-shape at Step~0, and that it is identical
with or without RoPE. Comparing initialized and pretrained networks,
we show that standard training does not overcome the topological valley, 
confirming that the U-shape persists as an architectural baseline 
under standard pretraining objectives.

We do not claim that this bias is insurmountable, nor that
interventions such as RoPE modifications are useless. We establish
what the baseline is and where it comes from, so that future efforts
to overcome it can be precisely targeted.
\end{abstract}

\section{Introduction}

As the context windows of Large Language Models (LLMs) have scaled from thousands to millions of tokens, a critical structural weakness has become apparent: models struggle to utilize information located in the middle of their input. This phenomenon manifests as a characteristic ``U-shaped'' accuracy curve, where retrieval and reasoning performance are high for tokens near the absolute beginning (primacy) and the absolute end (recency) of the prompt, but degrade severely in the middle \cite{liu2023lost, kazemnejad2024impact}.

While heavily studied empirically, the mechanistic cause of this phenomenon remains  debated. Current literature frequently attributes the Primacy effect (``Attention Sinks'') to a learned strategy where models dump excess Softmax probability mass on the first token \cite{xiao2023efficient}. Simultaneously, the middle-context degradation and recency bias are frequently blamed on the distance-decay properties of relative positional encodings like RoPE, sparking massive engineering efforts to flatten this decay \cite{su2024roformer, hsieh2024found}.

Concurrently, Herasimchyk et al. \cite{herasimchyk2026residual} developed a residual-aware theory using cumulative attention rollout to predict a U-shaped position bias. While their work is the closest to ours (see Section~\ref{sec:litreview}), their analysis relies on residual mixing coefficients measured from fully pretrained networks. Consequently, their claim of an ``architectural prior'' remains partially circular: by utilizing parameters derived from training, they cannot definitively decouple the structural topology from emergent, learned dynamics.

In this paper, we demonstrate the stronger, purely causal result: the U-shape exists at initialization, prior to any training whatsoever. We provide the first exact, closed-form mathematical proof that the U-shape is a fundamental topological constraint of the decoder-only architecture itself. By computing exact powers of the discrete Ces\`{a}ro matrix and deriving a closed-form continuous density in the large-$L$ limit, we obtain the exact equations governing gradient routing for random weights at \emph{Step 0}, establishing the U-shape as a true geometric birthright rather than an artifact of trained parameters or positional encodings.

We prove that causal masking algebraically guarantees a geometric Primacy bias, while residual connections guarantee a Recency bias. We empirically validate these equations by demonstrating that untrained, deep LLMs (such as a 24-layer Qwen2 architecture) exhibit a massive U-shaped gradient topology at \emph{Step 0}, prior to any training, and that this topology is mathematically identical regardless of the presence or absence of RoPE. 

Crucially, our continuous derivations reveal the precise physical mechanism that starves the middle context. While the final token can teleport its gradient directly via pure residual connections, and early tokens benefit from immense combinatorial compounding, intermediate tokens are forced to rely on \emph{hybrid paths}---trajectories that skip some layers via residuals but are subjected to the fractional dilution of the causal mixing matrices in others. This convolutional smearing creates a literal $\mathcal{O}(1/(H-1)!)$ dead zone in the middle of the context window. 

We emphasize that this structural dead zone is an architectural prior, not an insurmountable mathematical limit. During training, the optimizer attempts to overcome this structural bias by learning data-dependent, non-linear attention weights to route around the topological baseline. However, by deriving the exact closed-form calculus of the underlying linear geometry, we reveal the immense geometric headwinds the optimizer faces. Because standard pretraining objectives lack aggressive, targeted penalties to bridge this combinatorially suppressed valley, the model largely defaults to the path of least resistance: relying heavily on the geometric extremes.


\begin{framed}
\noindent\textbf{Scope.}\quad
We prove that the U-shape exists at initialization with random weights,
arises from causal masking and residual connections rather than
positional encodings, and persists under standard pretraining.
We do \emph{not} claim that it is insurmountable, that RoPE
modifications are useless, or that the Jacobian norm directly measures
retrieval accuracy. This paper derives the architectural baseline;
evaluating interventions to overcome it is future work.
\end{framed}

\section{Literature Review: The U-Shaped Attention Phenomenon}
\label{sec:litreview}

The empirical foundation for this work spans recent discoveries in LLM evaluation, attention analysis, and architectural inductive biases.

\paragraph{Empirical Discovery of the U-Shape.} 
Liu et al. (2023) \cite{liu2023lost} first formalized the ``Lost in the Middle'' phenomenon. By conducting multi-document QA tasks, they demonstrated that changing the absolute position of the relevant document within a long context drastically altered accuracy. Performance reliably followed a U-shaped curve, with GPT-3.5 and Claude exhibiting severe middle-context degradation. Subsequent studies confirmed this across various tasks, including legal analysis and code completion \cite{kazemnejad2024impact}.

\paragraph{Positional Attention Bias.}
Hsieh et al. (2024) \cite{hsieh2024found} traced the performance drop directly to the attention matrices. In ``Found in the Middle,'' they demonstrated that LLMs exhibit an intrinsic U-shaped \emph{positional attention bias} independent of content relevance. By calibrating the attention mechanism to subtract this purely positional component, they significantly improved middle-context utilization.

\paragraph{Graph-Theoretic and Causal Bias.}
The massive accumulation of attention on early tokens was first empirically formalized by Xiao et al.\ (2023) \cite{xiao2023efficient} as ``Attention Sinks,'' demonstrating that early tokens inevitably absorb massive probability mass simply due to their position at the root of the autoregressive chain. Seeking a structural explanation for this phenomenon, Wu et al.\ (2025) \cite{wu2025emergence} approached position bias from a graph-theoretic perspective. They proved that causal masking creates an asymmetric directed acyclic graph where early tokens lie on exponentially more computational paths than late tokens, driving a primacy bias that strengthens with depth. However, in their attention-only setting (no residual connections), their model predicts that cumulative attention collapses entirely onto the first token---predicting pure primacy rather than the empirically observed U-shape. This discrepancy was explicitly identified by the authors as an open problem. Both Herasimchyk et al.\ \cite{herasimchyk2026residual} and the present work independently resolve this contradiction by incorporating residual connections, which rescue the recency anchor and generate the complete topological U-shape.

\paragraph{Positional Encodings (RoPE) and Decay.}
The role of Rotary Position Embeddings (RoPE) \cite{su2024roformer} in context degradation has been widely studied. RoPE naturally enforces a long-term decay, heavily weighting local (recent) tokens. Work on scaling context windows (e.g., Ms-PoE) explicitly attempts to alter this distance-decay penalty to flatten the attention distribution and combat the recency bias \cite{su2024roformer, hsieh2024found}.

\paragraph{Residual-Aware Rollout.}
Herasimchyk et al. (2026) \cite{herasimchyk2026residual} developed a
residual-aware theory of cumulative attention rollout, incorporating
residual connections into cross-token information propagation. They
prove that causal masking induces primacy while residual connections
induce recency, and validate their predicted influence profiles against
pretrained models, achieving Spearman correlations of 0.88--0.98. Their
work is the most directly related to ours; the key distinction is that
we derive a closed-form density at initialization with random weights,
requiring no parameters from trained models.

\section{The Exact Calculus of Position Bias}
\label{sec:model}

To isolate the topological causes of the U-shape, we strip the transformer down to its routing components: causal attention and residual connections. Because position-wise feed-forward (MLP) networks apply identical feature transformations independently at each token position, their Jacobians are block-diagonal. They scale the magnitude of the signal but do not alter the macroscopic topology of the horizontal routing; thus, they are omitted from the topological model (see
Appendix~\ref{app:toy_model} for the full single-layer equation and
a precise account of which terms are retained, following the
linearisation approach of \citet{elhage2021mathematical}). We seek
to compute the Jacobian matrix of the final hidden state with respect to the input sequence, representing how much the input at position $j$ influences the final token $L$.

By demonstrating that the Score Pathway vanishes at initialization (see Appendix \ref{app:full_jacobian}), we can isolate the Value Pathway to calculate the network's inherent structural topology. While a fully trained network will inevitably utilize the non-linear Score Pathway to attempt to route around this baseline, isolating the Value Pathway provides the exact mathematical shape of the geometric headwind the optimizer must continuously fight against.

\subsection{The Discrete Causal Model (Ces\`{a}ro Matrices)}

At initialization, before the network has learned to correlate queries and keys, the expected dot product $q_i \cdot k_j \approx 0$. When passed through a causal Softmax, this yields a uniform distribution over all past tokens. We denote this base discrete causal attention matrix as $M$, where $M_{i,j} = 1/i$ for $j \le i$, and $0$ otherwise. 

This matrix is known in mathematics as the Ces\`{a}ro matrix. For a purely causal transformer without residuals, the gradient routed from the final token $L$ to an earlier token $j$ after $H$ layers is given by the bottom row of the exponential Ces\`{a}ro Matrix $M^H$. Using combinatorial identities, the exact discrete influence of token $j$ on token $L$ after $H$ causal layers is:

\begin{equation}
(M^H)_{L,j} \;=\; \binom{L-1}{j-1} \sum_{m=j}^{L} (-1)^{m-j} \binom{L-j}{m-j} \left(\frac{1}{m}\right)^H \label{eq:cesaro_discrete}
\end{equation}

When we introduce residual connections with a mixing weight $\alpha \in [0, 1]$, the layer update becomes the discrete residual matrix $N = (1-\alpha)I + \alpha M$. Expanding $N^H$ via the binomial theorem, the exact discrete influence equation becomes:

\begin{equation}
(N^H)_{L,j} \;=\; (1-\alpha)^H \,\delta_{L,j} \;+\; \sum_{r=1}^{H} \binom{H}{r} \alpha^r (1-\alpha)^{H-r} \left[ \binom{L-1}{j-1} \sum_{m=j}^{L} (-1)^{m-j} \binom{L-j}{m-j} \left(\frac{1}{m}\right)^r \right] \label{eq:recency_discrete}
\end{equation}

See Appendix \ref{app:cesaro-proof} for detailed proofs of the claims above.

\subsection{The Continuous Limit (Spherical-Cow Operator)}

To understand the macroscopic behavior of these exact discrete equations as sequence length $L \to \infty$, we map discrete positions $j \in \{1, \dots, L\}$ to a continuous coordinate $x = j/L \in (0, 1]$. We track the \emph{Influence Density} $\rho(x)$, defined as the input-output Jacobian norm representing how much the input at position $x$ influences the final hidden state at $x=1$. 

The discrete Ces\`{a}ro matrix $M$ converges to the linear continuous causal integral operator $\mathcal{M}$:
\begin{equation}
(\mathcal{M} h)(y) \;=\; \int_0^y \frac{1}{y}\, h(x)\, dx \label{eq:spherical_cow_cessaro}
\end{equation}
And the discrete residual matrix $N$ converges to the continuous residual operator $\mathcal{N}$:
\begin{equation}
(\mathcal{N} h)(y) \;=\; (1-\alpha)\, h(y) \;+\; \alpha (\mathcal{M} h)(y) \label{eq:spherical_cow_recency}
\end{equation}

By recursively evaluating the continuous kernel, we can decompose the position bias into its isolated structural ingredients.

\section{The Two Architectural Ingredients of the U-Shape}
\label{sec:ingredients}

We now unpack our exact topological solution into two distinct architectural properties that force the U-shape into existence at initialization, debunking the need for positional encodings (RoPE) to explain the right arm.

\subsection{Ingredient 1: Causal Masking (The Primacy Tail)}

If we temporarily disable residual connections ($\alpha=1$), the operator becomes pure causal averaging, $\mathcal{M}^H$. For a single layer ($H=1$), the continuous influence is exactly flat: $\rho_1^{(\mathcal{M})}(x) = 1$. There is no ``lost in the middle.''

However, for $H \ge 2$, early tokens become causally upstream of an exponentially growing number of integration paths. By recursively evaluating the uniform causal integral (see Appendix \ref{app:DetailedDerivations} for full proofs), we find the closed-form macroscopic continuous density:

\begin{equation}
\rho_H^{(\mathcal{M})}(x) \;=\; \frac{1}{(H-1)!} \left( \ln\frac{1}{x} \right)^{H-1} \label{eq:causal}
\end{equation}

As $x \to 0$, this function diverges. This logarithmic asymptote mathematically proves the Primacy effect (and the true origin of ``Attention Sinks,'' proving they are forced by geometry before they are utilized by the optimizer). Depth alone, combined with causal masking, forces influence to concentrate geometrically at the start of the prompt. This effect is a structural certainty of causal compounding.

\subsection{Ingredient 2: Residual Connections (The Recency Anchor)}

If we restore residual connections ($\alpha \in (0,1)$), we evaluate the influence of the full operator $\mathcal{N}^H$ at $x=1$  (see Appendix \ref{app:DetailedDerivations}  for details) yielding:

\begin{equation}
\rho_H^{(\mathcal{N})}(x) \;=\; (1-\alpha)^H \delta(1-x) \;+\; \sum_{r=1}^H \binom{H}{r} (1-\alpha)^{H-r} \alpha^r \frac{1}{(r-1)!}\left( \ln\frac{1}{x} \right)^{r-1} \label{eq:recency}
\end{equation}

This establishes an isolated Dirac delta spike at exactly $x=1$. The pure residual stream allows the final token to teleport its gradient directly to the output, establishing the necessary isolated anchor for the Recency effect without suffering the fractional dilution of passing through the causal mixing matrices.

Crucially, in deep networks (large $H$), the exponential factor $(r-1)!$ in the denominator severely squashes the magnitude of the continuous primacy tail relative to the $\mathcal{O}(1)$ residual highway of the final token. The characteristic U-shape emerges geometrically due to the interplay of these paths. Intermediate tokens lack the immense combinatorial compounding that rescues the earliest tokens. Instead, their primary signal arrives via hybrid paths—mathematically represented by the convolutions of the residual stream's Dirac deltas with the continuous causal mixing kernels. While these convolved hybrid paths smear the teleportation power across the context window (lifting the middle-context floor and preventing the gradient from collapsing to true zero), they still suffer from the fractional dilution of the integrations they do undergo. Thus, exact retrieval from the middle context remains structurally hostile under standard pretraining objectives.

The architectural extremes—the massive residual Recency Anchor on the right and the combinatorially compounded Primacy tail on the left—mathematically dwarf the diluted hybrid paths in the middle purely by virtue of the network's topology.

We note that multi-head attention does not alter this result; at
initialization it tightens the concentration around the theoretical
prediction (Appendix~\ref{app:multihead}).

\subsection{Why RoPE Is Irrelevant at Initialization}

Rotary Position Embeddings (RoPE) apply a position-dependent rotation
$R(\theta, j) \in O(2)$ to the query and key vectors before computing
the attention score:
\begin{equation}
    S_{ij} = \frac{(R_i\, q_i)^\top (R_j\, k_j)}{\sqrt{d_k}}
           = \frac{q_i^\top R(\theta, i-j)\, k_j}{\sqrt{d_k}}.
\end{equation}
At initialization, $q_i$ and $k_j$ are drawn independently from an
isotropic Gaussian. Because the distribution of an isotropic Gaussian
vector is invariant under orthogonal transformations, the rotated inner
product $q_i^\top R(\theta, i{-}j)\, k_j$ has the same distribution as
$q_i^\top k_j$ for any relative position $(i{-}j)$. Therefore RoPE
cannot break the uniformity of the expected attention distribution at
Step~0: the causal Softmax remains $A_{ij} \approx 1/i$ regardless of
whether RoPE is active.

This is not an approximation---it is an exact consequence of rotational
symmetry. The empirical confirmation (Spearman $\rho = 0.99$ between
RoPE and No-RoPE configurations at initialization;
Figure~\ref{fig:qwen2_init}) validates that finite-dimensional
deviations from perfect isotropy are negligible.

For absolute positional encodings (as in GPT-2), the analogous result
holds empirically: GPT-2 Small and Medium exhibit the same U-shape at
initialization (Appendix~\ref{app:generality}), though the theoretical
argument differs since absolute encodings break input isotropy by
additive shifts rather than rotations.

\section{Experimental Validation on Deep Architectures}
\label{sec:experiments}

The exact topological proof detailed above asserts that the U-shape is an architectural certainty of causal residuals, present at initialization (Step 0), independent of RoPE. Furthermore, our theory predicts that the shape is inherently asymmetric: the Primacy tail decays logarithmically, while the Recency anchor is an isolated $\mathcal O(1)$ Dirac delta spike. 

To empirically validate our continuous differential calculus against real, untrained decoder-only networks, we initialize a standard deep architecture (Qwen2-0.5B: $H=24$ layers, 896-dimension, SwiGLU, RMSNorm) with random Gaussian weights. We calculate the exact empirical Input-Output Jacobian norm across the sequence length $L=2048$, isolating the architectural biases prior to any data-dependent learning.

Combining the scalar surrogate from Appendix~\ref{app:experiments} with
the reduction in Appendix~\ref{app:full_jacobian}, the quantity plotted
on a logarithmic scale is
\begin{equation}
    \log_{10}\rho(j) \;=\; H\log_{10}\gamma \;+\; \log_{10}(N^H)_{L,j},
\end{equation}
where $\gamma = \|W_V W_O\|$ is a position-invariant per-layer gain.
Since the first term is constant across positions, the \emph{shape} of
the plotted curve is governed exclusively by the Ces\`{a}ro kernel
$(N^H)_{L,j}$.

\begin{figure}[htbp]
\centering
\includegraphics[width=\textwidth]{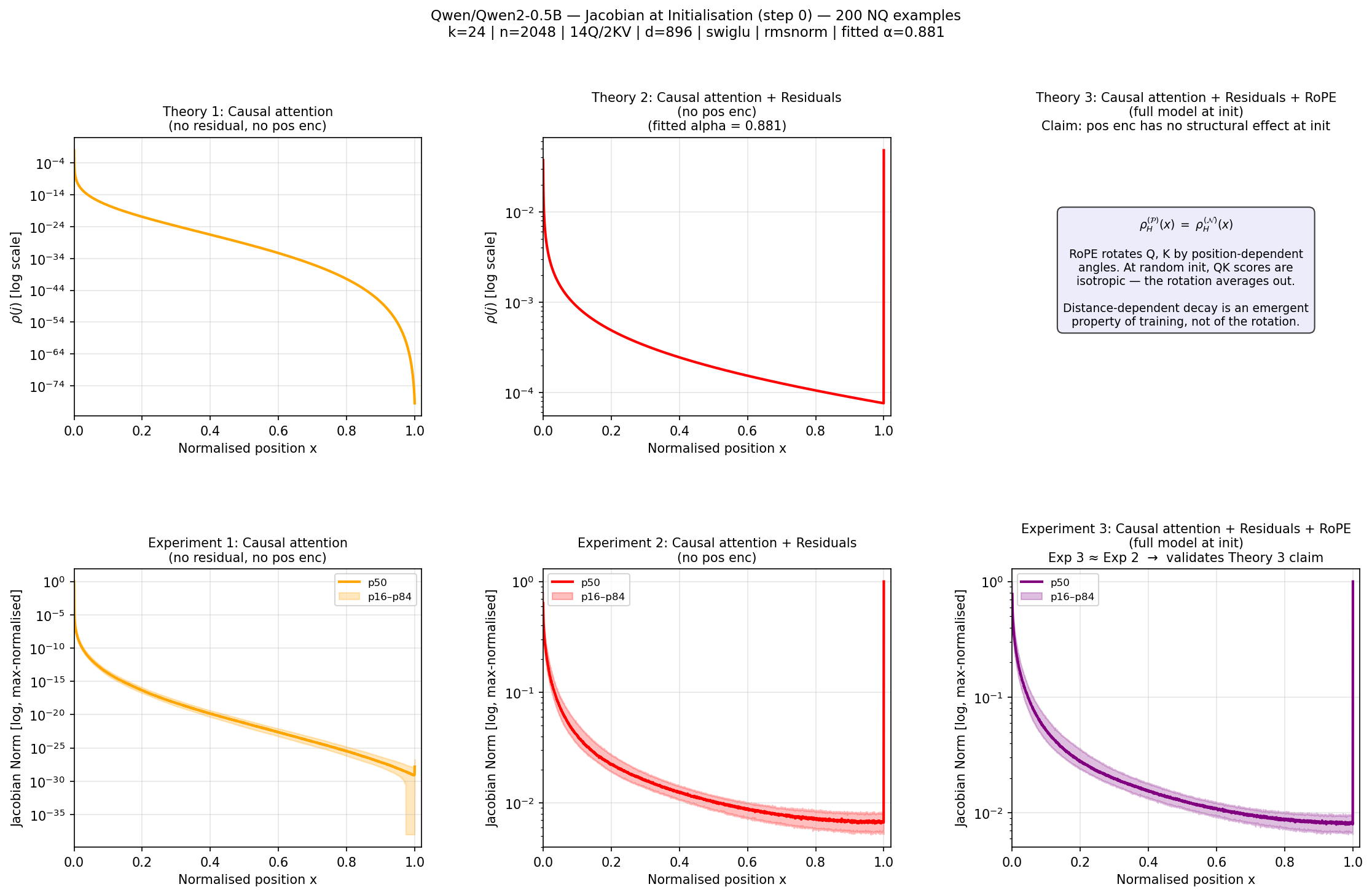}
\caption{\textbf{Empirical Validation of Position Bias on Qwen2-0.5B at Initialization (Step 0).} 
\emph{Top Row (Theory):} The exact topological solutions for pure Causal Primacy (Eq. \ref{eq:causal}), Hybrid Residual Recency (Eq. \ref{eq:recency}) (plotted on a logarithmic scale), and the theoretical irrelevance of positional encodings at initialization.
\emph{Bottom Row (Empirical):} The measured Input-Output Jacobian norm $\rho(x)$ for the 24-layer Qwen2 architecture prior to training. The empirical network tightly recovers the mathematically derived asymmetric U-shape. Crucially, Exp 3 demonstrates that Rotary Positional Embeddings (RoPE) have no structural effect on the topological U-shape at initialization, confirming our hypothesis that middle-context degradation is an inherent property of causal residuals.}
\label{fig:qwen2_init}
\end{figure}

As shown in Figure \ref{fig:qwen2_init}, our exact continuous theory tightly predicts the massive asymmetric U-shape. To rigorously evaluate the goodness-of-fit between our closed-form analytical theory (Eq. \ref{eq:causal} and Eq. \ref{eq:recency}) and the empirical Step 0 Jacobian, we calculate the Spearman rank correlation and the Wasserstein distance. Our continuous equations achieve a Spearman correlation of $\rho = 0.99$ and a Wasserstein distance of $\mathcal{W} = 0.02$, confirming that our idealized Ces\`{a}ro operator accurately captures the discrete network topology. We further validate the universality of this topology by demonstrating identical U-shaped behavior in GPT-2 architectures (see Appendix~\ref{app:generality}).

Crucially, empirical measurement of the Qwen2 Jacobian identically confirms this theory. Removing RoPE entirely from the 24-layer Qwen2 network at initialization produces an exact topological match to the fully-functional RoPE network (Spearman $\rho = 0.99$ between RoPE and No-RoPE configurations): massive causal compounding at $x=0$, a dead zone in the middle, and a massive $\mathcal O(1)$ gradient teleportation spike via the residual connections at exactly $x=1$.

The widespread engineering effort in the literature to flatten relative positional encodings (e.g., LongRoPE, YaRN, ALiBi) fundamentally misunderstands the geometry of the network. The middle context falls into the mathematical gap between geometric path expansion (left) and residual teleportation (right), governed exclusively by matrix depth $H$ and fractional Softmax dilution.

\section{Initialization vs. Pretraining: The Rigid Topological Baseline}
\label{sec:pretraining}

To demonstrate that this mathematically derived U-shape serves as the
architectural baseline of the Transformer, we compare the
exact empirical Input-Output Jacobian of a 24-layer Qwen2-0.5B
architecture at initialization (random weights) against the exact same
architecture fully pretrained on billions of tokens. We average the
Jacobian norm across 200 distinct sequences from the NaturalQuestions
(NQ) multi-document QA dataset \cite{kwiatkowski2019natural}, each formatted to a
context window of $L=2048$ tokens.

To test whether the pretrained model's Jacobian spikes reflect learned
detection of content discontinuities, we additionally evaluate on
\emph{chunked} sequences: 300-token excerpts from distinct NQ documents
concatenated with no separator tokens, with chunk boundaries aligned at
fixed positions (0, 300, 600, \ldots) across all 200 sequences (full
details in Appendix~\ref{app:evolution}). If the spikes are learned
boundary detectors, they should appear at regular 300-token intervals
in the chunked condition for the pretrained model but not at
initialization.

\begin{figure}[htbp]
\centering
\includegraphics[width=\textwidth]{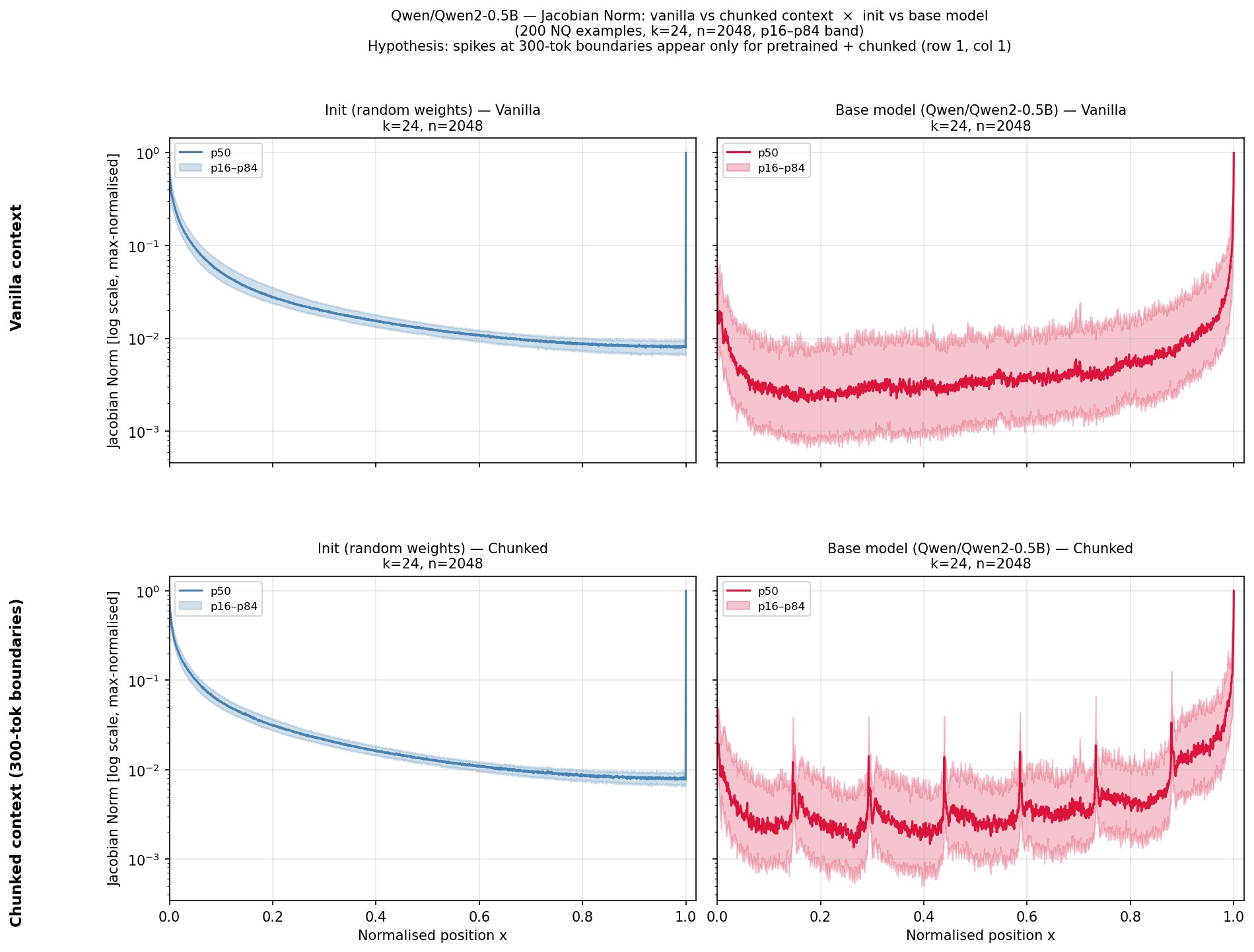}
\caption{\textbf{Jacobian Norm: Initialization vs.\ Pretrained Qwen2-0.5B
($H=24$, $L=2048$), Vanilla vs.\ Chunked Context.}
Averaged over 200 NQ sequences with p16--p84 percentile bands.
\emph{Top row (Vanilla):} Each sequence is a single NQ document truncated
to 2048 tokens.
\emph{Bottom row (Chunked):} Each sequence concatenates 300-token excerpts
from distinct NQ documents with no separator tokens, boundaries aligned at
positions 0, 300, 600, \ldots\
\emph{Left column (Initialization):} Both conditions exhibit the smooth
Ces\`{a}ro U-shape with no sensitivity to content or chunk boundaries.
\emph{Right column (Pretrained):} The macroscopic U-shape persists. In the
chunked condition, sharp spikes emerge at the 300-token document
boundaries, confirming that these are learned content-discontinuity
detectors rather than positional artifacts.}
\label{fig:init_vs_pretrained}
\end{figure}

As shown in Figure~\ref{fig:init_vs_pretrained}, the initialized
network closely obeys the pure mathematical architecture in both
conditions: a smooth, asymmetric topological geometry---the continuous
causal Primacy tail and the discrete $\mathcal O(1)$ residual Recency spike---with
no learned routing and no sensitivity to chunk boundaries. The pretrained
model (right column) preserves the macroscopic U-shape but introduces
sharp localised spikes. In the chunked condition (bottom row), these
spikes align with the 300-token document boundaries, confirming that they
represent learned structural routing to content discontinuities rather
than fixed positional artifacts.

Note that the U-shape is not confined to the global context window. Any sub-interval bounded by attention anchors---chunk boundaries, formatting delimiters, or segment markers---exhibits its own local U-shape governed by the same Ces\`{a}ro kernel (Figures~\ref{fig:init_vs_pretrained}). The architectural prior is \emph{scale-free}: it re-instantiates at every level of the sequence hierarchy wherever causal masking operates between boundary tokens.

During training on highly structured data, the model attempts to overcome this baseline. The right panel demonstrates how the model's learned routing (the ``flesh'') is forced to conform to the underlying architectural geometry (the ``bone structure''). The model learns to ``smear'' the recency anchor backward across the final 10-15\% of tokens to capture the immediate question context, and it learns to create localized spikes for important syntactic markers (e.g., document boundaries or prompt formatting). 

However, despite this learned task-specific routing, the model does not flatten the macroscopic $\mathcal{O}(1/(H-1)!)$ geometric dead zone in the middle of the context window. The topological baseline drops so severely due to the fractional dilution of the causal mask that overcoming it would require an aggressive, targeted loss penalty that standard next-token prediction lacks. Thus, the model defaults to the path of least resistance: relying heavily on the geometric extremes.

This analysis demonstrates that the ``Lost in the Middle'' phenomenon is fundamentally constrained by an architectural baseline that standard pretraining does not overcome. Training attempts to mitigate this baseline by creating localized attention spikes to grab useful information, but the sheer geometric depth of the causal-residual valley makes exact retrieval from the middle context structurally hostile under standard objectives. On a logarithmic scale, the peak-to-trough ratio is approximately $10^2$ at initialization and $10^3$  after pretraining, confirming that standard training does not compress the topological valley. This persistence reflects a fundamental asymmetry in the backward pass:
the gradient used to \emph{learn} to attend to middle-context positions
is attenuated by the same positional factor as the forward signal. The
optimizer receives an effective per-position learning rate
$\eta(x) \propto \rho_H(x)$: positions in the dead zone update
factorially more slowly than the extremes, regardless of the global
learning rate.

To visualise how the Score Pathway activates during early training while
the Value Pathway baseline holds rigid, we train a freshly initialised
Qwen2-0.5B for 100 gradient steps on NQ data and measure the Jacobian
on 10 held-out vanilla sequences at each step (full methodology in
Appendix~\ref{app:evolution}). As shown in
Figure~\ref{fig:jacobian_evolution}, the Step~0 curve (dark blue)
exhibits the smooth Ces\`{a}ro shape predicted by our theory. Within
approximately 50 steps, sharp spikes emerge as the Score Pathway
activates, but the macroscopic U-shaped envelope persists underneath.
Crucially, the middle-context floor never rises---on the normalised
(right) panel, the relative depth of the valley actually increases
during training, confirming that the optimizer defaults to the geometric
path of least resistance rather than bridging the combinatorially
suppressed dead zone.

\begin{figure}[htbp]
\centering
\includegraphics[width=\textwidth]{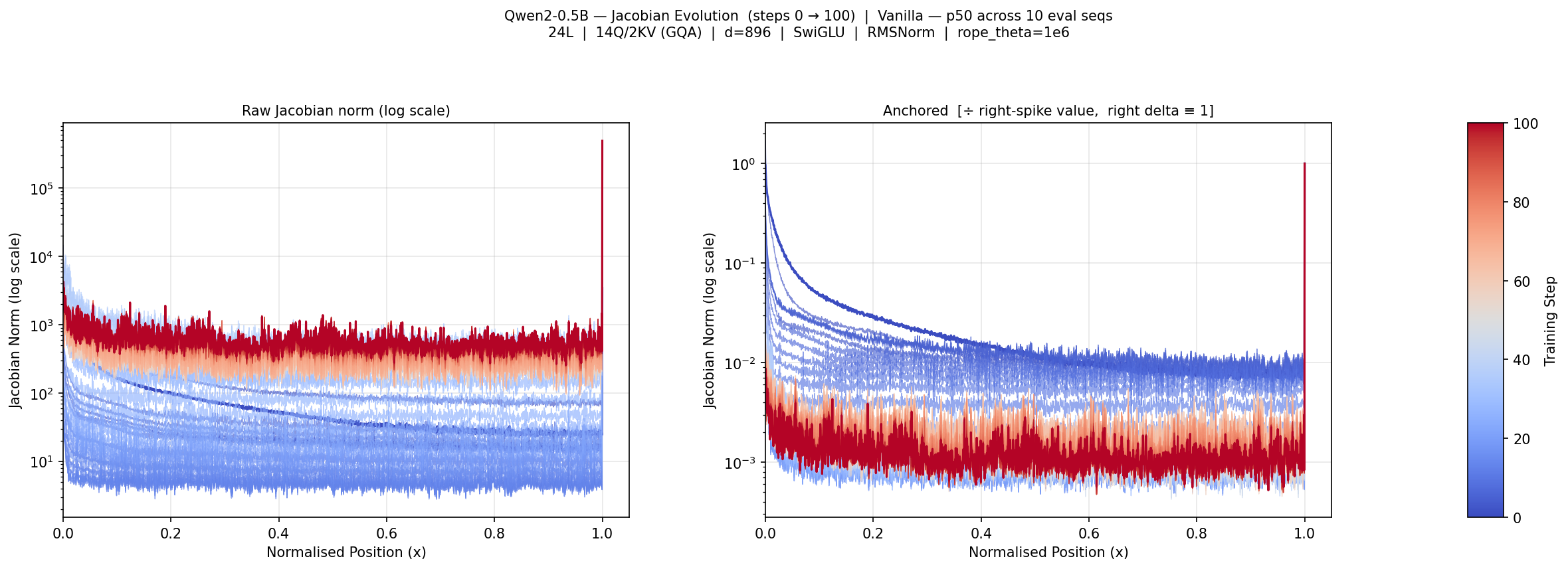}
\caption{\textbf{Evolution of the Jacobian Topology During Early Pretraining (Steps 0--100).} 
Measured on Qwen2-0.5B with a sequence length of $L=2048$.
\emph{Left (Raw Magnitude):} The global gradient norm decreases as the model stabilizes, but the macroscopic U-shaped topology persists rigidly. 
\emph{Right (Anchored):} By normalizing the Recency Anchor ($x=1$) to $1.0$, we observe that the relative depth of the ``Lost-in-the-Middle'' valley actually \emph{increases} during training. The optimizer does not flatten the combinatorially suppressed middle section, instead relying increasingly on the geometric path of least resistance: the residual Recency Anchor and the logarithmic Primacy Tail.}
\label{fig:jacobian_evolution}
\end{figure}

\section{Conclusion and Future Work}
We have presented an exact analytical model proving that the ``Lost in the Middle'' phenomenon is a structural inevitability of deep autoregressive transformers. Causal masking guarantees primacy; residuals guarantee a recency anchor.

Our findings suggest a paradigm shift in how the community addresses context degradation. Because the U-shape is a topological birthright rather than a positional encoding artifact, architectural tweaks like flattening RoPE merely treat the symptom. To cure the disease, training paradigms must be explicitly designed to overcome the $\mathcal{O}(1/(H-1)!)$ initialization bias.

A critical direction for future work is to evaluate whether specialized training paradigms---such as explicit middle-context curriculum learning, targeted loss weighting, or over-sampling ``needle-in-a-haystack'' data---can force the non-linear Score Pathway to completely bridge this topological gap. By providing the exact closed-form calculus of the causal-residual baseline, this paper equips future researchers with the precise physical headwinds their optimization strategies must overcome.

Future empirical work should focus on directly measuring the Input-Output Jacobian norm across long context windows in open-source models (e.g., Llama 3) to verify the precise logarithmic scaling of the primacy tail and the exponential scaling of the recency smear derived in this paper.

\section*{Limitations}

Our theoretical derivations cleanly isolate the linear Value Pathway by demonstrating the Score Pathway vanishes at initialization. However, our analysis of fully trained networks relies on empirical observation of the Jacobian rather than closed-form mathematical bounds on the trained Softmax. Furthermore, our empirical validation focuses exclusively on standard pretraining distributions. Determining the upper bound of the Score Pathway's ability to override the topological baseline under aggressive, position-targeted fine-tuning remains an open empirical question.

\section*{Disclaimer}

BC worked on this project in a personal capacity. The views and conclusions expressed are solely those of the authors and do not reflect those of Meta. This work is independent of BC’s professional duties and is neither affiliated with nor endorsed by Meta.

\section*{Acknowledgments and AI Disclosure}
The author acknowledges the use of Anthropic Claude (via Claude Code) and Perplexity AI as coding and writing assistants during the preparation of this manuscript. These tools were utilized to assist with \LaTeX\ typesetting, grammatical proofreading, and writing Python scripts for statistical analysis, model inference, and the 100-step micro-training experiment. All theoretical frameworks, mathematical proofs, experimental designs, and data interpretations are the original contributions of the author.

\appendix

\section{Proof of the Exact Ces\`{a}ro Matrix Power Entry}
\label{app:cesaro-proof}

\textbf{Theorem.} 
Let $M \in \mathbb{R}^{L \times L}$ be the lower--triangular Ces\`{a}ro matrix
\[
M_{i,j} \;=\;
\begin{cases}
\dfrac{1}{i}, & 1 \le j \le i \le L,\\[4pt]
0, & 1 \le i < j \le L.
\end{cases}
\]
Then for any $H \ge 1$,
\begin{equation}
(M^H)_{i,j}
=\begin{cases}
\binom{i-1}{j-1}
\sum_{m=j}^{i}
(-1)^{\,m-j}\binom{i-j}{m-j}\left(\frac{1}{m}\right)^{H},
& 1 \le j \le i \le L,\\[8pt]
0, & 1 \le i < j \le L.
\end{cases}
\label{eq:cesaro-power-general}
\end{equation}
In particular, taking $i=L$ yields \eqref{eq:cesaro_discrete} in the main text.

\textbf{Proof.} 
The upper-triangular case $(i<j)$ is immediate: since $M$ is strictly lower-triangular, $M^H$ is strictly lower-triangular for all $H \ge 1$.

For the lower-triangular case $(j \le i)$, we proceed by induction on $H$.

\paragraph{Base case ($H=1$).}
For $H=1$ and $j \le i$, the right-hand side becomes
\[
\binom{i-1}{j-1}
\sum_{m=j}^{i}
(-1)^{\,m-j}\binom{i-j}{m-j}\frac{1}{m}.
\]
With the change of variables $k = m-j$ this is
\[
\binom{i-1}{j-1}
\sum_{k=0}^{i-j}
(-1)^k \binom{i-j}{k}\frac{1}{k+j}.
\]
Writing $\frac{1}{k+j} =\int_0^1 x^{k+j-1} dx$, the inner sum is the integral representation of the Beta function
\[
\sum_{k=0}^{i-j}
(-1)^k \binom{i-j}{k}\frac{1}{k+j}
= B\bigl(j,\,i-j+1\bigr)
= \frac{1}{i\binom{i-1}{j-1}}.
\]
Multiplying by $\binom{i-1}{j-1}$ gives
\[
(M^1)_{i,j} = \begin{cases}
\dfrac{1}{i}, & 1 \le j \le i \le L,\\[4pt]
0, & 1 \le i < j \le L.
\end{cases},
\]
which is exactly $M_{i,j}$ by definition. The base case holds.

\paragraph{Inductive step.}
Assume \eqref{eq:cesaro-power-general} holds for some $H \ge 1$.
Then for $H+1$ and $j \le i$,
\[
(M^{H+1})_{i,j}
= \sum_{k=j}^{i} M_{i,k}(M^H)_{k,j}
= \frac{1}{i}\sum_{k=j}^{i}(M^H)_{k,j},
\]
since $M_{i,k}=1/i$ for all $j \le k \le i$.

Substituting the inductive hypothesis for  $(M^H)_{k,j}$,
\[
(M^{H+1})_{i,j}
= \frac{1}{i}\sum_{k=j}^{i}
\left[
\binom{k-1}{j-1}
\sum_{m=j}^{k}
(-1)^{\,m-j}\binom{k-j}{m-j}\left(\frac{1}{m}\right)^{H}
\right].
\]
We now swap the order of summation over the region $j \le m \le k \le i$:
\[
(M^{H+1})_{i,j}
= \frac{1}{i}\sum_{m=j}^{i}
(-1)^{\,m-j}\left(\frac{1}{m}\right)^{H}
\left[
\sum_{k=m}^{i}
\binom{k-1}{j-1}\binom{k-j}{m-j}
\right].
\]

The inner sum simplifies via the binomial identity
\[
\binom{k-1}{j-1}\binom{k-j}{m-j}
= \binom{m-1}{j-1}\binom{k-1}{m-1},
\]
so that
\[
\sum_{k=m}^{i}
\binom{k-1}{j-1}\binom{k-j}{m-j}
= \binom{m-1}{j-1}\sum_{k=m}^{i}\binom{k-1}{m-1}.
\]
Applying the Hockey-Stick identity
$\sum_{k=m}^{i}\binom{k-1}{m-1} = \binom{i}{m}$,
we obtain
\[
\sum_{k=m}^{i}
\binom{k-1}{j-1}\binom{k-j}{m-j}
= \binom{m-1}{j-1}\binom{i}{m}.
\]

Substituting back,
\[
(M^{H+1})_{i,j}
= \frac{1}{i}\sum_{m=j}^{i}
(-1)^{\,m-j}\left(\frac{1}{m}\right)^{H}
\binom{m-1}{j-1}\binom{i}{m}.
\]
A straightforward factorial simplification yields
\[
\frac{1}{i}\binom{i}{m}\binom{m-1}{j-1}
= \frac{1}{m}\binom{i-1}{j-1}\binom{i-j}{m-j}.
\]
Thus,
\[
(M^{H+1})_{i,j} 
= \begin{cases}  \binom{i-1}{j-1}
\sum_{m=j}^{i}
(-1)^{\,m-j}\binom{i-j}{m-j}
\left(\frac{1}{m}\right)^{H+1}, & 1 \le j \le i \le L,\\[4pt]
0, & 1 \le i < j \le L.
\end{cases} \label{eq:M}
\]
which is exactly \eqref{eq:cesaro-power-general} with $H$ replaced by $H+1$.
By induction on $H$, the formula holds for all $H \ge 1$ and $1 \le j \le i \le L$.
\qed

\section{ Integral Representation of $(M^H)_{i,j}$}

To simplify $(M^H)_{i,j}$ we use the identity for the power function via the Gamma function:
\begin{equation}
\frac{1}{m^H} = \frac{1}{(H-1)!} \int_0^\infty t^{H-1} e^{-mt} dt
\end{equation}
Substituting this into the summation and invoking the \textit{Fubini-Tonelli theorem} to interchange the order of summation and integration, we obtain:
\begin{equation}
(M^H)_{i,j} = \frac{1}{(H-1)!} \binom{i-1}{j-1} \int_0^\infty t^{H-1} \left[ \sum_{m=j}^{i} (-1)^{m-j} \binom{i-j}{m-j} e^{-mt} \right] dt
\end{equation}
Letting $k = m - j$, the inner sum simplifies via the Binomial Theorem:
\begin{equation}
e^{-jt} \sum_{k=0}^{i-j} \binom{i-j}{k} (-e^{-t})^k = e^{-jt} (1 - e^{-t})^{i-j}
\end{equation}
Applying the change of variables $u = e^{-t}$ (where $dt = -du/u$), the integral transforms to:
\begin{equation}
(M^H)_{i,j} =  \frac{1}{(H-1)!} \binom{i-1}{j-1} \int_0^1 (-\ln u)^{H-1} u^{j-1} (1-u)^{i-j} du
\end{equation}

\section{Detailed Derivations of Density Operators in the Continuum Limit}
\label{app:DetailedDerivations}

In the continuous limit $L \to \infty$, vectors become continuous functions $h(x)$, and matrix multiplication becomes an integral operator:
\begin{equation}
(\mathcal{O} h)(y) \;=\; \int_0^1 G_1(y,x)\, h(x)\, dx
\end{equation}
where $G_1(y,x)$ is the \emph{integral kernel}. For the linear operator applied $n$ times, one obtains:
\begin{equation}
(\mathcal{O}^n h)(y) \;=\; \int_0^1 G_n(y,x)\, h(x)\, dx
\end{equation}
where mathematical consistency demands the convolution relation:
\begin{equation}
G_{m+n}(y,x) \;=\; \int_0^1 G_m(y,z)\, G_n(z,x)\, dz
\end{equation}

In the context of autoregressive language modeling, we are specifically concerned with predicting the next token, which depends entirely on the hidden state of the final token of the prompt. We define the \emph{influence density} $\rho_n(x)$ as the kernel evaluated at this final token ($y=1$):
\begin{equation}
    \rho_n(x) \;\equiv\; G_n(1, x)
\end{equation}

We now derive the exact closed-form influence density for both the pure causal kernel and the causal kernel with residual connections.

\subsection{Influence Density for the Pure Causal Kernel}

Let $K_n(y,x)$ denote the $n$-layer kernel for the pure causal attention operator $\mathcal{M}$. If there are no layers ($n=0$), the output is identical to the input, yielding the identity kernel:
\begin{equation}
K_0(y,x) \;=\; \delta(y-x)
\end{equation}
The first non-trivial operator applies a uniform average over all past tokens:
\begin{equation}
    (\mathcal{M} h)(y) \;=\; \int_0^y \frac{1}{y}\, h(x)\, dx
\end{equation}
By factoring in the causal mask using the Heaviside step function $\Theta(y-x)$, we can extract the layer-1 kernel directly:
\begin{equation}
K_1(y,x) \;=\; \frac{1}{y} \Theta(y-x)
\end{equation}

We will prove by mathematical induction that the general $n$-layer causal kernel takes the closed form:
\begin{equation}
K_n(y,x) \;=\; \frac{1}{y} \frac{1}{(n-1)!} \left( \ln\frac{y}{x} \right)^{n-1} \Theta(y-x)
\end{equation}

\paragraph{Proof.} 
For the base case $n=1$, the formula yields $\frac{1}{y} \frac{1}{0!} (\ln\frac{y}{x})^0 \Theta(y-x) = \frac{1}{y} \Theta(y-x)$, which trivially holds.

Assume the formula holds for layer $n$. To find the kernel for layer $n+1$, we apply the consistency convolution between $K_1$ and $K_n$:
\begin{equation}
K_{n+1}(y,x) \;=\; \int_0^1 K_1(y,z)\, K_n(z,x)\, dz
\end{equation}
Substituting our known expressions:
\begin{equation}
K_{n+1}(y,x) \;=\; \int_0^1 \left[ \frac{1}{y} \Theta(y-z) \right] \left[ \frac{1}{z} \frac{1}{(n-1)!} \left(\ln\frac{z}{x}\right)^{n-1} \Theta(z-x) \right] dz
\end{equation}
The product of the Heaviside functions $\Theta(y-z)\Theta(z-x)$ restricts the region of non-zero integration strictly to $x \le z \le y$. If $x > y$, the integral vanishes, preserving the global causal mask $\Theta(y-x)$. For $x \le y$, we extract the constants with respect to $z$:
\begin{equation}
K_{n+1}(y,x) \;=\; \frac{1}{y} \frac{1}{(n-1)!} \int_x^y \frac{1}{z} \left(\ln\frac{z}{x}\right)^{n-1} dz
\end{equation}
We evaluate this integral using integration by substitution. Let $u = \ln(z/x)$, which implies $du = \frac{1}{z} dz$. The integration bounds transform from $z \in [x, y]$ to $u \in [0, \ln(y/x)]$:
\begin{equation}
K_{n+1}(y,x) \;=\; \frac{1}{y} \frac{1}{(n-1)!} \int_0^{\ln(y/x)} u^{n-1}\, du
\end{equation}
\begin{equation}
K_{n+1}(y,x) \;=\; \frac{1}{y} \frac{1}{(n-1)!} \left[ \frac{u^n}{n} \right]_0^{\ln(y/x)} \;=\; \frac{1}{y} \frac{1}{n!} \left(\ln\frac{y}{x}\right)^n
\end{equation}
Re-attaching the global causal mask $\Theta(y-x)$ completes the inductive step. $\qed$

\vspace{1em}
Finally, we evaluate this kernel at the final token $y=1$ to recover the macroscopic influence density of the pure causal architecture. Because the normalized sequence coordinate $x \in (0, 1]$, the condition $x \le 1$ is universally satisfied, meaning $\Theta(1-x) = 1$. Thus, the primacy tail density after $n$ layers is exactly:
\begin{equation}
\rho_n^{(M)}(x) \;=\; K_n(1,x) \;=\; \frac{1}{(n-1)!} \left( \ln\frac{1}{x} \right)^{n-1}
\end{equation}
This rigorously establishes the logarithmic divergence at the start of the prompt caused by pure causal compounding.

\subsection{Influence Density for the Causal Residual Kernel}

Let $\tilde K_n(y,x)$ denote the $n$-layer kernel for the network with both causal attention and residual connections. For a zero-layer architecture, the causal residual kernel is trivially identical to the pure causal identity kernel:
\begin{equation}
\tilde K_0(y,x) \;=\; K_0(y,x) \;=\; \delta(y-x)
\end{equation}
The single-layer operator applies a weighted average of a residual connection and the uniform causal average over all past tokens:
\begin{equation}
(\mathcal{N} h)(y) \;=\; (1-\alpha) h(y) + \alpha \int_0^y \frac{1}{y} h(x)\, dx
\end{equation}
This gives us the single-layer kernel:
\begin{align}
\tilde K_1(y,x) \;&=\; (1-\alpha) \delta(y-x)  + \alpha \frac{1}{y} \Theta(y-x) \\
\;&=\; (1-\alpha) K_0(y,x) + \alpha K_1(y,x) 
\end{align}

We will prove by mathematical induction that the general $n$-layer residual kernel is a binomial expansion of the pure causal kernels:
\begin{equation}
\tilde K_n(y,x) \;=\; \sum_{k=0}^n \binom{n}{k} \alpha^k (1-\alpha)^{n-k} K_k(y,x)
\end{equation}

\paragraph{Proof.}
The base case $n=1$ yields $\binom{1}{0} \alpha^0 (1-\alpha)^1 K_0 + \binom{1}{1} \alpha^1 (1-\alpha)^0 K_1 = (1-\alpha)K_0 + \alpha K_1$, which trivially holds.

Assume the formula holds for layer $n$. To find the kernel for layer $n+1$, we apply the consistency convolution $\tilde K_{n+1}(y,x) = \int_0^1 \tilde K_1(y,z) \tilde K_n(z,x)\, dz$:
\begin{equation}
\tilde K_{n+1} \;=\; \int_0^1 \Big[ (1-\alpha) K_0(y,z) + \alpha K_1(y,z) \Big] \left[ \sum_{k=0}^n \binom{n}{k} \alpha^k (1-\alpha)^{n-k} K_k(z,x) \right] dz
\end{equation}
Because $K_0(y,z) = \delta(y-z)$, its convolution simply extracts the $n$-layer kernel. For $K_1$, we rely on the consistency relation proved in the previous section: $\int_0^1 K_1(y,z) K_k(z,x) dz = K_{k+1}(y,x)$. Distributing the integral gives:
\begin{equation}
\tilde K_{n+1} \;=\; (1-\alpha) \sum_{k=0}^n \binom{n}{k} \alpha^k (1-\alpha)^{n-k} K_k + \alpha \sum_{k=0}^n \binom{n}{k} \alpha^k (1-\alpha)^{n-k} K_{k+1}
\end{equation}
To combine these sums, we re-index the second sum by letting $j = k+1$, and we substitute $j=k$ in the first sum:
\begin{equation}
\tilde K_{n+1} \;=\; \sum_{j=0}^n \binom{n}{j} \alpha^j (1-\alpha)^{n+1-j} K_j + \sum_{j=1}^{n+1} \binom{n}{j-1} \alpha^j (1-\alpha)^{n+1-j} K_j
\end{equation}
We group the terms by $K_j$. For $j=0$ and $j=n+1$, the boundary terms map perfectly to $\binom{n+1}{0}$ and $\binom{n+1}{n+1}$. For $1 \le j \le n$, we factor out the common coefficients:
\begin{equation}
\left[ \binom{n}{j} + \binom{n}{j-1} \right] \alpha^j (1-\alpha)^{n+1-j} K_j
\end{equation}
By Pascal's Rule, $\binom{n}{j} + \binom{n}{j-1} = \binom{n+1}{j}$. Substituting this identity confirms that the combined sum is exactly:
\begin{equation}
\tilde K_{n+1}(y,x) \;=\; \sum_{j=0}^{n+1} \binom{n+1}{j} \alpha^j (1-\alpha)^{n+1-j} K_j(y,x)
\end{equation}
This completes the inductive step. $\qed$

\vspace{1em}
To find the final influence density for the residual network, we evaluate this full kernel at $y=1$. Separating the purely residual $k=0$ term from the causal paths $k \ge 1$, and substituting our closed-form $K_k(1, x)$ expressions, we obtain the exact architectural baseline:
\begin{equation}
\rho_n^{(\mathcal{N})}(x) \;=\; \underbrace{(1-\alpha)^n \delta(1-x)}_{\text{Recency Anchor}} \;+\; \underbrace{\sum_{k=1}^n \binom{n}{k} \alpha^k (1-\alpha)^{n-k} \frac{1}{(k-1)!} \left( \ln\frac{1}{x} \right)^{k-1}}_{\text{Continuous Causal \& Hybrid Paths}}
\end{equation}
This derivation proves that causal mixing inherently creates the logarithmic primacy tail, but the residual stream bypasses the mixing matrix entirely ($k=0$), resulting in an infinite Dirac density spike exactly at $x=1$.

\section{From Full Transformer Layer to the Ces\`{a}ro Toy Model}
\label{app:toy_model}

This appendix makes precise which components of a full transformer
layer are retained in the positional routing model of
Section~\ref{sec:model}, following the linearisation philosophy of
\citet{elhage2021mathematical}.

\subsection*{Full Layer}

A single pre-norm transformer block maps $h^{(0)}_i$ to $h^{(1)}_i$ via:
\begin{equation}
    h^{(1)}_i = h^{(0)}_i
    + \tilde{\alpha}\,(M\,g(h^{(0)}))_i
    + f\!\Big(g\!\big(h^{(0)}_i
      + \tilde{\alpha}\,(M\,g(h^{(0)}))_i\big)\Big)
    \label{eq:full_layer}
\end{equation}
where $M_{ij} = \frac{1}{i}\,\Theta(i-j)$ is the causal averaging
(Ces\`{a}ro) matrix, $g = \mathrm{RMSNorm}$,
$f(x) = \mathrm{GELU}(x\,W_1)\,W_2$ is the feed-forward network,
and $\tilde{\alpha} = \|W_V W_O\|$ is the gain of the value-output
pathway (denoted $\gamma$ in Appendix~\ref{app:full_jacobian}).
Only $M$ mixes information across token positions; $f$ and $g$
act per-token.

\subsection*{Simplifications}

We make two simplifications to isolate the positional routing:

\begin{equation}
    h^{(1)}_i =
    \underbrace{h^{(0)}_i\vphantom{\Big(}}_{\text{keep}}
    + \underbrace{\tilde{\alpha}\vphantom{\Big(}}_{\text{keep}}
    \underbrace{(\,M\vphantom{\Big(}}_{\text{keep}}
    \underbrace{g(h^{(0)})\,)_i}_{\substack{\text{drop } g:\\ g \to \mathrm{Id}}}
    + \underbrace{f\!\Big(g\!\big(h^{(0)}_i
      + \tilde{\alpha}(Mg(h^{(0)}))_i\big)\Big)}_{\text{drop entirely}}
\end{equation}

\begin{enumerate}
    \item \textbf{Drop $g$ (RMSNorm).}
    At initialisation, RMSNorm rescales all tokens to the same RMS,
    contributing a position-independent per-token gain. Removing it
    preserves the positional routing structure.

    \item \textbf{Drop $f$ (FFN).}
    The feed-forward network acts per-token and does not directly
    introduce cross-position mixing. However, its input
    $h^{(0.5)}_i = h^{(0)}_i + \tilde{\alpha}(Mg(h^{(0)}))_i$
    already contains the attention output, so the nonlinearity in
    $f$ couples to the positional mixing from $M$.
    Section~\ref{sec:model} notes that the FFN Jacobian is
    block-diagonal and does not alter the topology of horizontal
    routing; dropping $f$ removes the nonlinear coupling while
    preserving this property.
\end{enumerate}

The resulting toy model is:
\begin{equation}
    h^{(1)} = (I + \tilde{\alpha}\,M)\,h^{(0)}
\end{equation}

After $H$ layers:
\begin{equation}
    h^{(H)} = (I + \tilde{\alpha}\,M)^H\,h^{(0)}
\end{equation}

\subsection*{Normalised Form}

The rows of $(I + \tilde{\alpha}\,M)^H$ sum to
$(1 + \tilde{\alpha})^H$, reflecting the fact that the residual
connection adds signal rather than redistributing it. For the
purpose of studying the \emph{shape} of the influence profile, we
normalise by defining
\begin{equation}
    \alpha = \frac{\tilde{\alpha}}{1 + \tilde{\alpha}},
\end{equation}
so that
\begin{equation}
    N = (1-\alpha)\,I + \alpha\,M
\end{equation}
is a stochastic matrix whose rows sum to one, and $(N^H)_{L,j}$ is
a probability distribution over source positions. This is the form
used throughout the paper. The physical gain
$\tilde{\alpha} = \gamma = \|W_V W_O\|$ relates to the fitted
parameter via $\tilde{\alpha} = \alpha/(1-\alpha)$.
Note that $\gamma^H$ factors out as a position-uniform constant
(Appendix~\ref{app:full_jacobian}), so the shape of the U-curve
depends only on $\alpha$, not on $\gamma$ separately.

\section{Full Single-Head Jacobian and Reduction to the Ces\`{a}ro Kernel}
\label{app:full_jacobian}

This appendix derives the complete back-propagation Jacobian of a single
causal attention layer, identifies precisely the roles of $W_Q$, $W_K$,
$W_V$, and $W_O$, and shows the exact approximation step that reduces the
position-routing component to the scalar Ces\`{a}ro kernel used throughout
the paper.

\subsection{Setup}

Consider a single attention head with input sequence
$\mathbf{H}^{(l)} \in \mathbb{R}^{L \times d}$.
The standard attention mechanism computes:
\begin{align}
    Q &= \mathbf{H}^{(l)} W_Q, \quad
    K = \mathbf{H}^{(l)} W_K, \quad
    V = \mathbf{H}^{(l)} W_V,\\
    S_{ij} &= \frac{q_i \cdot k_j}{\sqrt{d_k}} + m_{ij},
    \qquad m_{ij} = -\infty \text{ for } j > i,\\
    A_{i,\cdot} &= \mathrm{softmax}(S_{i,\cdot}) \in \mathbb{R}^{i},\\
    \mathbf{z}_i &= \sum_{j \le i} A_{ij}\, v_j,
    \qquad
    \mathbf{o}_i = \mathbf{z}_i W_O.
\end{align}
With a residual connection, the layer update is
$\mathbf{h}^{(l+1)}_i = \mathbf{h}^{(l)}_i + \mathbf{o}_i$.

\subsection{Full Per-Layer Jacobian}

The Jacobian of the \emph{output} at position $i$ with respect to the
\emph{input} at position $j \le i$ is:
\begin{equation}
    \frac{\partial \mathbf{h}^{(l+1)}_i}{\partial \mathbf{h}^{(l)}_j}
    \;=\;
    \delta_{ij}\, I_d
    \;+\;
    \underbrace{A_{ij}\, W_V W_O}_{\displaystyle\text{Value Pathway}}
    \;+\;
    \underbrace{
      \left(\sum_{k \le i}
      \frac{\partial A_{ik}}{\partial \mathbf{h}^{(l)}_j}\,
      v_k \right) W_O
    }_{\displaystyle\text{Score Pathway}}.
\label{eq:full_jacobian}
\end{equation}

\paragraph{Value Pathway.}
Perturbing $\mathbf{h}^{(l)}_j$ changes $v_j = \mathbf{h}^{(l)}_j W_V$
directly. This perturbation is mixed into $\mathbf{z}_i$ with weight
$A_{ij}$, then projected by $W_O$. This pathway is active for all
$j \le i$ and is the direct, ``vertical'' route through $W_V$.

\paragraph{Score Pathway.}
Perturbing $\mathbf{h}^{(l)}_j$ also changes $q_j$ and $k_j$, modifying
the attention scores $S_{ij}$, and hence the routing weights $A_{ij}$.
Using the chain rule through the causal Softmax:
\begin{equation}
    \frac{\partial A_{ik}}{\partial \mathbf{h}^{(l)}_j}
    \;=\;
    \sum_{m \le i}
    A_{ik}\bigl(\delta_{km} - A_{im}\bigr)
    \left[
      \delta_{ij}\,\frac{W_Q\, k_m^\top}{\sqrt{d_k}}
      \;+\;
      \delta_{mj}\,\frac{q_i\, W_K^\top}{\sqrt{d_k}}
    \right].
\label{eq:score_path}
\end{equation}
The two Kronecker deltas correspond to two distinct mechanisms:
\begin{itemize}
    \item $\delta_{ij}$: the query $q_i$ changes (only when $j = i$,
    i.e.\ a token attending to itself).
    \item $\delta_{mj}$: the key $k_j$ changes, modifying how much every
    later token attends to position $j$.
\end{itemize}

\subsection{Why the Score Pathway Vanishes at initialization}

At random Gaussian initialization, $W_Q$ and $W_K$ are zero-mean with
entries $\mathcal O(d_k^{-1/2})$ (standard Kaiming or Xavier scaling). Consequently:
\begin{enumerate}
    \item The pre-softmax scores satisfy
    $S_{ij} = q_i \cdot k_j / \sqrt{d_k} \;\xrightarrow{d_k\to\infty}\; 0$,
    and so the causal Softmax is approximately uniform:
    $A_{ij} \approx 1/i$ for all $j \le i$.
    \item The Softmax Jacobian at a uniform distribution evaluates to
    $A_{ik}(\delta_{km} - A_{im}) = (1/i)(\delta_{km} - 1/i)$,
    which is $\mathcal O(1/i)$.
    \item Each term in the Score Pathway (Eq.\ \ref{eq:score_path}) carries
    an additional factor of $W_Q k_m^\top$ or $q_i W_K^\top$, both
    $\mathcal O(d_k^{-1/2})$, which suppresses the Score Pathway by
    $\mathcal O(d_k^{-1/2})$ relative to the Value Pathway.
\end{enumerate}

In the standard wide-network regime ($d_k \to \infty$), the Score Pathway
contribution vanishes at initialization \cite{poole2016exponential,
yang2019scaling}. This is the formal justification for the Softmax
linearisation used in the main text.

\subsection{Reduction to the Scalar Ces\`{a}ro Kernel}

Retaining only the Value Pathway, Eq.~\ref{eq:full_jacobian} simplifies to:
\begin{equation}
    \frac{\partial \mathbf{h}^{(l+1)}_i}{\partial \mathbf{h}^{(l)}_j}
    \;\approx\;
    \delta_{ij}\, I_d
    \;+\;
    A_{ij}\, W_V W_O.
\end{equation}
The positional structure is governed entirely by the scalar $A_{ij}$, while
$W_V W_O \in \mathbb{R}^{d \times d}$ is a position-\emph{invariant} feature
mixing matrix, identical for all token pairs $(i, j)$.

The end-to-end Jacobian after $H$ layers is obtained by chaining these
terms. Taking norms and factoring out the position-independent ``vertical''
gain $\gamma = \|W_V W_O\|$ per layer:
\begin{equation}
    \rho(j) \;=\;
    \left\| \frac{\partial \mathbf{h}^{(H)}_L}{\partial \mathbf{h}^{(0)}_j}
    \right\|
    \;\approx\;
    \gamma^H \cdot (N^H)_{L,j},
\end{equation}
where $N = (1-\alpha)I + \alpha M$ is the residual Ces\`{a}ro matrix and
$(N^H)_{L,j}$ is its $(L,j)$ entry. Since $\gamma^H$ is a constant that
shifts all positions uniformly on the log scale, it has no effect on the
\emph{shape} of the U-curve.

\paragraph{Conclusion.}
$W_V$ does not generate the kernel: it provides a uniform scalar gain that
is identical at every position. The asymmetric U-shape is an exclusive
property of the scalar routing matrix $M^H$ (or $N^H$), determined solely
by the causal mask and the residual mixing parameter $\alpha$.

\subsection{Multi-Head Attention at Initialization}
\label{app:multihead}

At initialization, all heads share the same uniform attention
distribution $A_{ij} = 1/i$, so the expected positional structure is
identical to the single-head Ces\`{a}ro kernel. However, each head
contributes an independent random gain $W_V^h W_O^h$, so the total
Jacobian is a sum of $H_{\text{heads}}$ independent terms with
identical positional structure. By concentration of measure, the
empirical Jacobian norm converges to the theoretical prediction as
$\mathcal{O}(1/\sqrt{H_{\text{heads}}})$. Increasing the number of
heads therefore \emph{tightens} the architectural prior: the U-shape
becomes less variable across random initializations, not more.

\section{What is Plotted: The Scalar Surrogate for the Jacobian Norm}
\label{app:experiments}

We measure how strongly each input position $j$ influences the model's \emph{final} prediction at the last position $L$. Let the input embedding matrix be $E \in \mathbb{R}^{L \times d}$, with $e_j \in \mathbb{R}^d$ representing the embedding vector at position $j$. Let the final-layer hidden state at the last position be $h_L^{(H)} \in \mathbb{R}^d$. We define the logit vector (pre-softmax) at the last position as:
\begin{equation}
\ell \;=\; W_U h_L^{(H)} \in \mathbb{R}^{|V|}
\end{equation}
where $W_U \in \mathbb{R}^{|V|\times d}$ is the unembedding matrix.

The full input-output Jacobian of the last-position logits with respect to the input embeddings is the tensor:
\begin{equation}
J \;=\; \frac{\partial \ell}{\partial E} \in \mathbb{R}^{|V| \times L \times d}
\end{equation}
We define the per-position slice of this Jacobian as $J_j \equiv \frac{\partial \ell}{\partial e_j} \in \mathbb{R}^{|V|\times d}$.

A mathematically natural scalar ``influence score'' per position would be the true Frobenius norm of this slice:
\begin{equation}
\rho_F(j) \;=\; \left\|J_j\right\|_F \;=\; \sqrt{\sum_{v=1}^{|V|}\sum_{k=1}^{d}\left(\frac{\partial \ell_v}{\partial e_{j,k}}\right)^2}
\end{equation}
However, computing $\|J_j\|_F$ exactly is computationally expensive, as it requires aggregating gradient information from all $|V|$ independent logits (effectively requiring $|V|$ separate backward passes). 

Instead, our experimental code utilizes a highly efficient one-pass \emph{scalar-probe} surrogate. We choose a fixed probe vector $u \in \mathbb{R}^{|V|}$ (in our implementation, the all-ones vector $u=\mathbf{1}$), and define the scalar surrogate:
\begin{equation}
s \;=\; u^\top \ell = \sum_{v=1}^{|V|} \ell_v
\end{equation}
A single backward pass from this scalar $s$ yields the $d$-dimensional gradient vector:
\begin{equation}
g_j \;\equiv\; \frac{\partial s}{\partial e_j}
\;=\; \left(\frac{\partial \ell}{\partial e_j}\right)^\top u
\;=\; J_j^\top u
\;\in\; \mathbb{R}^{d}
\end{equation}
The quantity we plot is the standard $L_2$ (Euclidean) norm of this vector:
\begin{equation}
\rho(j) \;\equiv\; \|g_j\|_2 \;=\; \sqrt{\sum_{k=1}^d (g_{j,k})^2}
\end{equation}

Importantly, $\rho(j)=\|J_j^\top u\|_2$ is \emph{not} identically equal to the Frobenius norm $\|J_j\|_F$. Taking the Euclidean norm of a sum of vectors is not equivalent to the square root of the sum of their squared norms. 

However, in the structural regime studied in this paper, this surrogate perfectly preserves the macroscopic topology. The position-dependent structure of the network is dominated exclusively by the scalar routing kernel $(N^H)_{L,j}$, while the remaining feature mixing (including the accumulated products of $W_V W_O$ and $W_U$) behaves as a position-invariant matrix $C \in \mathbb{R}^{|V| \times d}$. Because we can factor $J_j \approx (N^H)_{L,j} \cdot C$, both the Frobenius norm and our scalar surrogate scale linearly with the structural kernel:
\begin{align}
    \rho_F(j) \;&\propto\; |(N^H)_{L,j}| \cdot \|C\|_F \\
    \rho(j) \;&\propto\; |(N^H)_{L,j}| \cdot \|C^\top u\|_2 
\end{align}

Because the $L_2$ norm behaves dimensionally identical to the square root of a sum of squares (analogous to RMSE, not MSE), taking the logarithm separates the structural kernel from the position-invariant constants cleanly. Concretely, the quantity displayed in our figures is plotted on a logarithmic scale as:
\begin{equation}
\log_{10}\rho(j) \;=\; H\log_{10}\gamma \;+\; \log_{10}(N^H)_{L,j}
\end{equation}
where $\gamma$ collects the position-invariant constants from the feature-mixing matrices. Thus, the positional \emph{shape} of the empirical U-curve is governed exclusively by $(N^H)_{L,j}$, confirming that our scalar surrogate rigorously captures the exact geometric topology of the network.

\section{Training Evolution Experiment: Setup and Methodology}
\label{app:evolution}

To observe how the architectural prior evolves during early training, we
conduct a controlled micro-training experiment on a freshly initialised
Qwen2-0.5B architecture (494M parameters, 24 layers, 14Q/2KV GQA,
$d=896$, SwiGLU, RMSNorm, $\theta_{\text{rope}}=10^6$, sequence length
$L=2048$). No pretrained weights are loaded.

\paragraph{Data.}
We stream the NaturalQuestions (NQ) validation set and buffer the first
60 examples. Examples 0--49 form the training pool; examples 50--59 are
held out for evaluation and are never seen during training.

\paragraph{Training.}
The model is trained for 100 steps using AdamW
($\mathrm{lr}=5\times10^{-4}$ decayed to $5\times10^{-5}$ via cosine
schedule, weight decay $0.1$, gradient clipping at $1.0$, batch size 1).
At each step, a random contiguous window of 2048 tokens is drawn from a
randomly selected training-pool sequence and a standard next-token
cross-entropy loss is computed.

\paragraph{Eval sequences.}
We construct two parallel evaluation sets from the 10 held-out examples:
\begin{itemize}
    \item \textbf{Vanilla.} Each held-out NQ example is tokenised as a
    single document (question concatenated with document tokens),
    truncated to 2048 tokens.
    \item \textbf{Chunked.} Each eval sequence is constructed by
    concatenating 300-token excerpts drawn from different
    \emph{training-pool} documents. A per-sequence offset cycles through
    the pool so that content varies across sequences, but chunk
    boundaries are aligned at fixed token positions
    (0, 300, 600, \ldots) across all 10 sequences. No separator or
    delimiter tokens are inserted between chunks. This alignment ensures
    that any boundary effects are preserved---rather than washed
    out---when averaging Jacobians across sequences.
\end{itemize}

\paragraph{Jacobian measurement.}
Before each gradient update (and once after the final step), we compute
the $\ell_2$ norm of the input-embedding Jacobian (as defined in
Appendix~\ref{app:experiments}) for all 10 vanilla and 10 chunked eval
sequences independently. This produces two arrays of shape
$(10, 101, 2048)$, recording the full evolution of the Jacobian norm
profile from Step~0 through Step~100.

\section{Generality of the Phenomenon Across Architectures}
\label{app:generality}

To demonstrate that the asymmetric U-shape is a universal geometric property invariant to depth or specific architectural choices (e.g., RoPE, SwiGLU, RMSNorm), we provide the corresponding initialization and pretraining Jacobian comparisons for the classic GPT-2 series (GPT-2 Small, $H=12$; GPT-2 Medium, $H=24$). 

\begin{figure}[htbp]
\centering
\begin{minipage}{.48\textwidth}
  \centering
  \includegraphics[width=\linewidth]{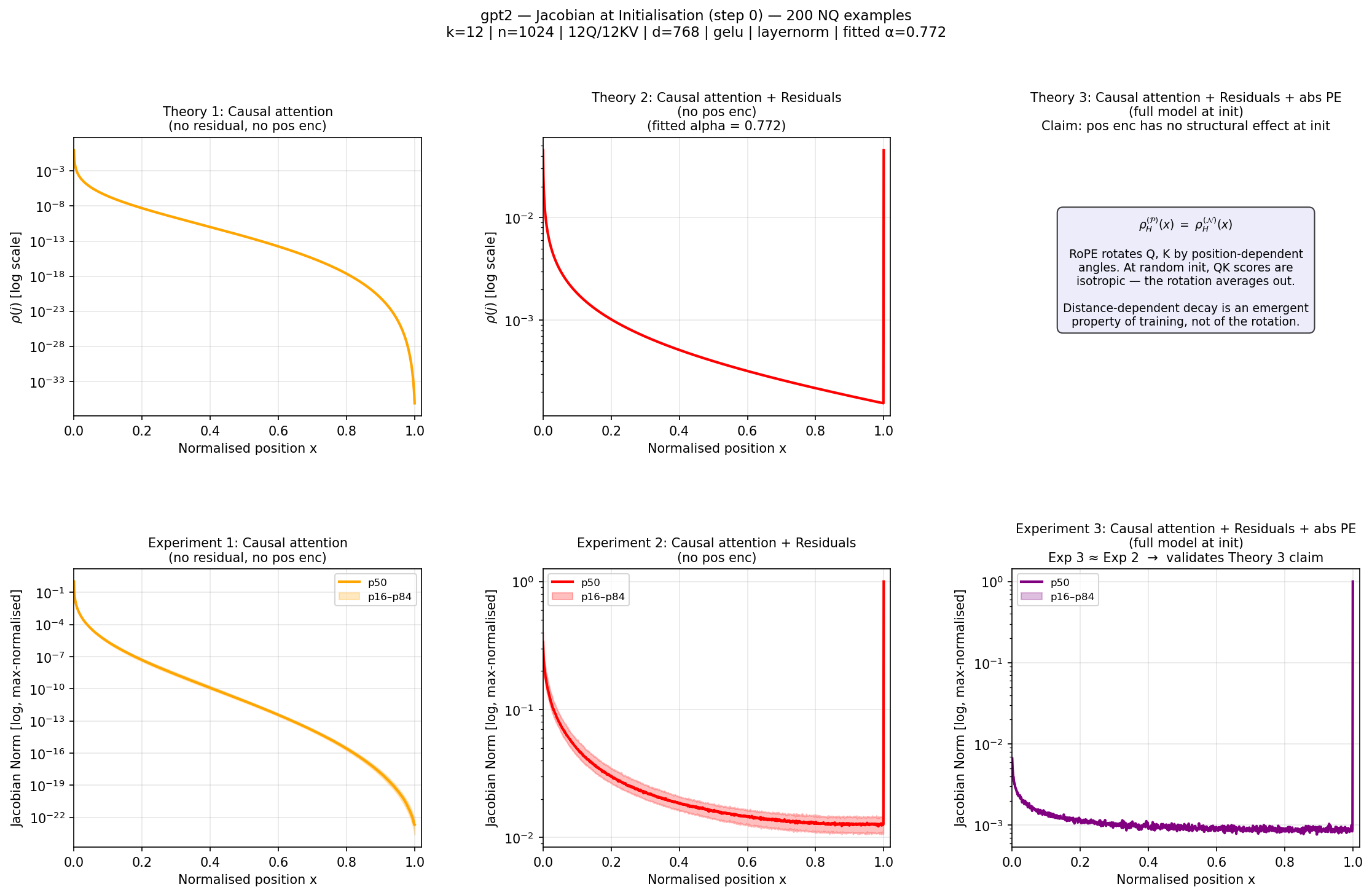}
  \caption{Init: GPT-2 Small ($H=12$)}
  \label{fig:gpt2_initA}
\end{minipage}%
\hfill
\begin{minipage}{.48\textwidth}
  \centering
  \includegraphics[width=\linewidth]{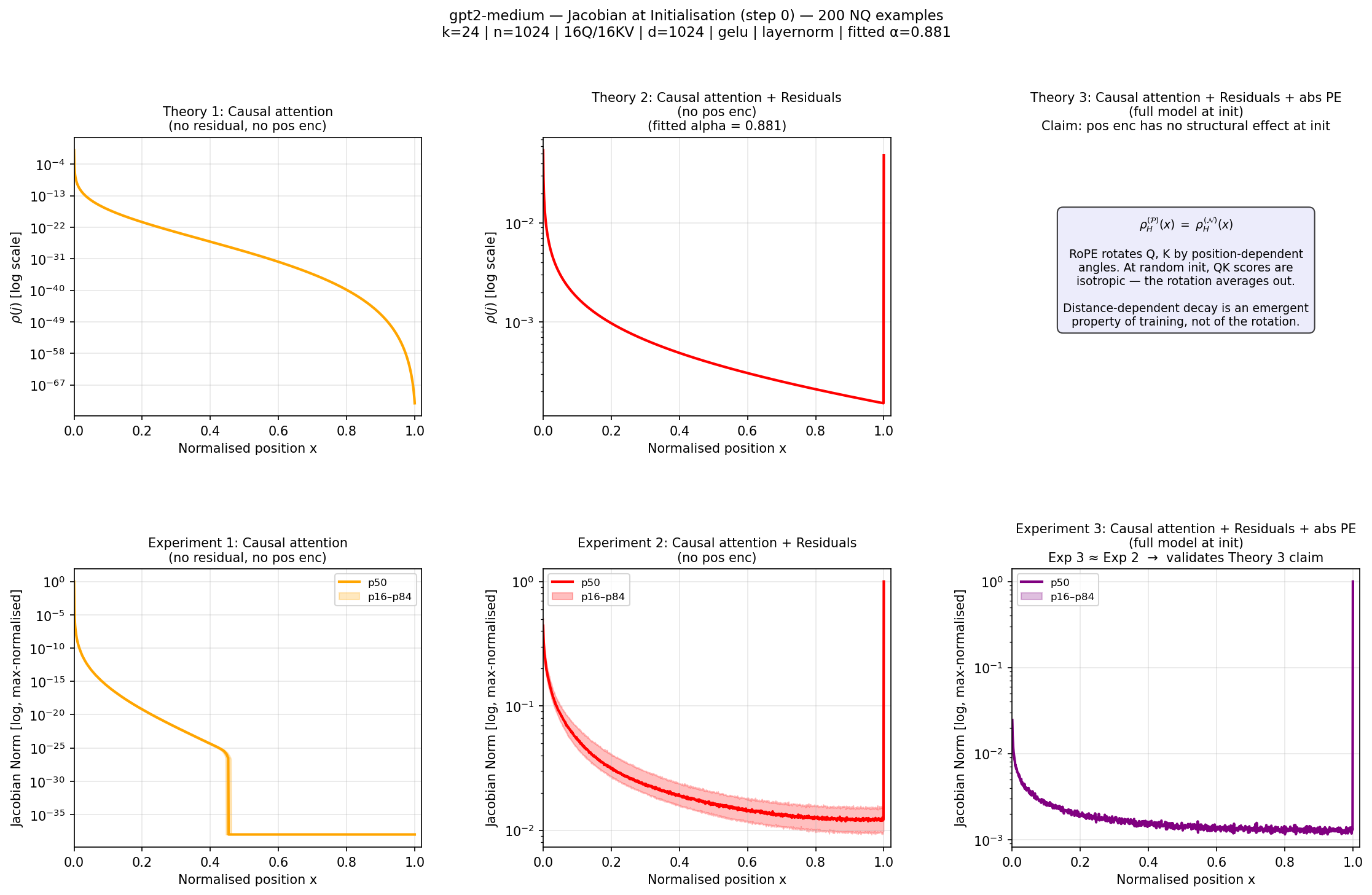}
  \caption{Init: GPT-2 Medium ($H=24$)}
  \label{fig:gpt2_initB}
\end{minipage}
\end{figure}

\begin{figure}[htbp]
\centering
\begin{minipage}{.48\textwidth}
  \centering
  \includegraphics[width=\linewidth]{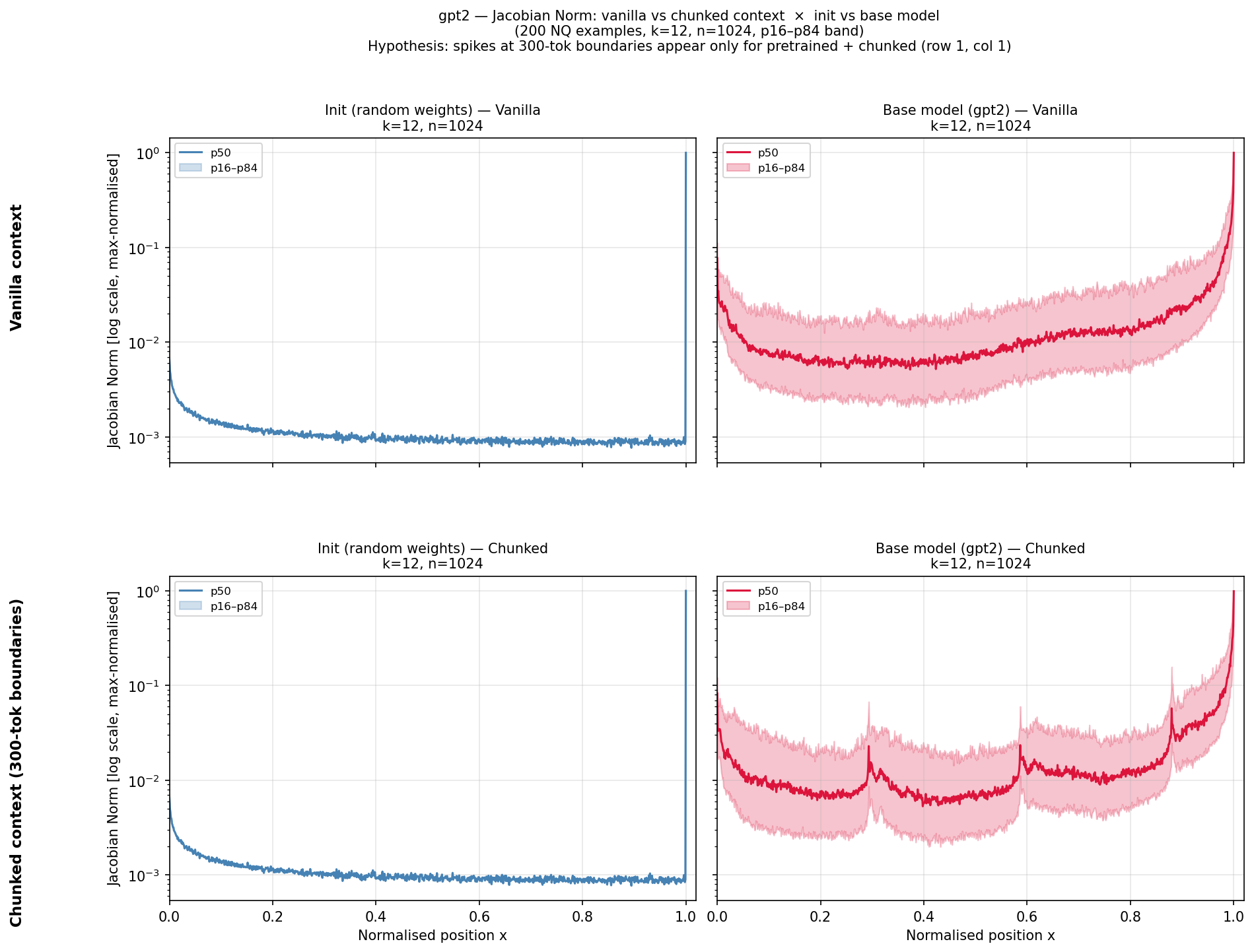}
  \caption{Pretrained vs. Init: GPT-2 Small}
  \label{fig:gpt2_pretrainedA}
\end{minipage}%
\hfill
\begin{minipage}{.48\textwidth}
  \centering
  \includegraphics[width=\linewidth]{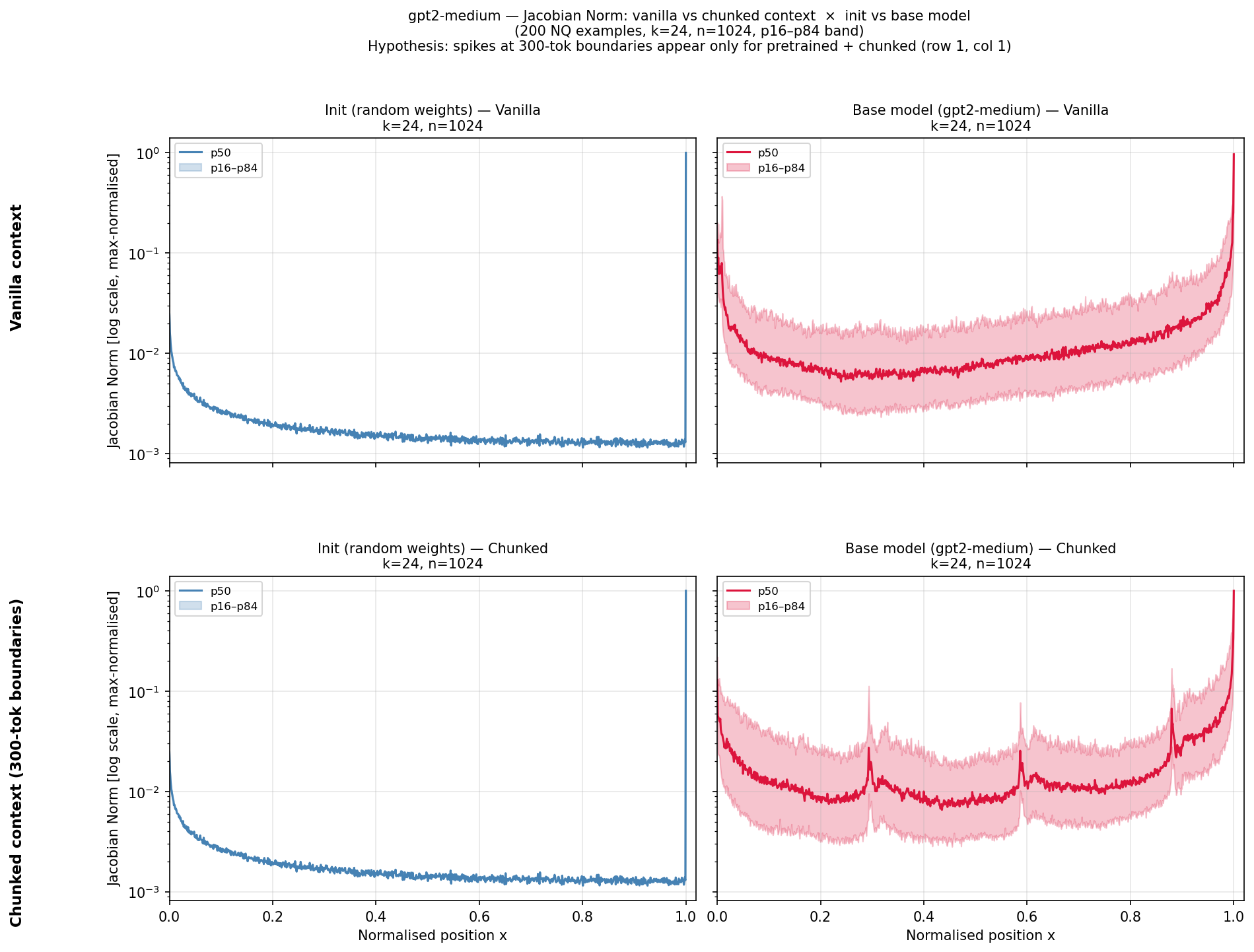}
  \caption{Pretrained vs. Init: GPT-2 Medium}
  \label{fig:gpt2_pretrainedB}
\end{minipage}
\end{figure}

As shown in Figures \ref{fig:gpt2_initA}, \ref{fig:gpt2_initB}, \ref{fig:gpt2_pretrainedA} and \ref{fig:gpt2_pretrainedB}, models utilizing standard LayerNorm, absolute positional encodings, and Gelu activations exhibit identical baseline topologies to the modern Qwen2 architecture shown in Figures \ref{fig:qwen2_init} and \ref{fig:init_vs_pretrained}. In all cases, the topological U-shape dominates at initialization, and its macroscopic constraints persist rigidly through pretraining.

\bibliography{references}
\end{document}